# 3D reconstruction through fusion of cross-view images

Rongjun Qin[1,2,*], Shuang Song[1], Xiao Ling[1], Mostafa Elhashash[2]

1. Department of Civil, Environmental and Geodetic Engineering, The Ohio State University, United States, 2036 Neil Avenue, Columbus, Ohio, USA. qin.324@osu.edu

2. Department of Electrical and Computer Engineering, The Ohio State University, United States.
* Corresponding author

## Abstract

3D recovery from multi-stereo and stereo images, as an important application of the image-based perspective geometry, serves many applications in computer vision, remote sensing and Geomatics. In this chapter, the authors utilize the imaging geometry and present approaches that perform 3D reconstruction from cross-view images that are drastically different in their viewpoints. We introduce our framework that takes ground-view images and satellite images for full 3D recovery, which includes necessary methods in satellite and ground-based point cloud generation from images, 3D data co-registration, fusion and mesh generation. We demonstrate our proposed framework on a dataset consisting of twelve satellite images and 150k video frames acquired through a vehicle-mounted Go-pro camera and demonstrate the reconstruction results. We have also compared our results with results generated from an intuitive processing pipeline that involves typical geo-registration and meshing methods.

## I. Introduction

3D data generation often requires expensive data collection such as aerial photogrammetric or LiDAR flight [1, 2]. Depending on the required accuracy, resolution and other specs of the final products, the efforts in data collection and processing can exponentially grow. Alternative and low-cost data sources are of particular interest for wide-area 3D modeling [3]: Satellite sensors running 24/7 offer overview images covering large regions with single scans, which comparatively come with lower cost than aerial flights and do not require physical access to the area of interest [4]. On the other hand, there exist many ground-view images coming either from crowdsourcing platforms or collected using relatively low-cost equipment (e.g. video frames from low-cost cameras) that provides high-resolution information of objects. Both the overview and the ground-view data are complementary to each other and their view differences being approximately 90° forms cross-view dataset, a fusion of which may yield a low-cost solution for city-scale 3D modeling. This Chapter describes our ongoing work (an earlier work is described in [5]) in an attempt to address this challenging task by proposing an integrated framework to fuse the 3D results of satellite overview and ground-view video frames to generate 3D textured mesh models presenting both top and side view features.

The available commercial satellite images often have 0.3 - 0.5 meter GSD (ground sampling distance) and ground-view images can easily reach a GSD of a few millimeters. With significantly different resolution, the resulting 3D geometry may be associated with different uncertainties, which adds additional challenges for the fusion task of these two types of data, which include:

1) The quality of 3D output separately generated from satellite images and ground-view images are scene-specific and may differ in terms of completeness and accuracy. Algorithms and basic principles for addressing image-based 3D modeling are relative standard, thus the image quality and their respective characteristics play a major role in the reconstruction results, such as the photo-consistency / temporal differences / illumination among images, their geometric setup, completeness in terms of coverage, and intersection angles etc.

2) Due to the large view differences, the overview and ground-view dataset may share very limited region in common, and additionally the 3D output from the ground-view dataset may come with no geo-referencing information and may contain non-rigid topographic distortions (e.g. trajectories drift or distortions due to inaccurate interior/exterior orientation estimation), which further add challenges in 3D geo-registration of the dataset.

3) The combined 3D point clouds are from two sources with different resolution, uncertainty and radiometric properties of textures, which present difficulties in both the geometric reconstruction of meshes and the texture mappings. Thus, obtaining visually consistent textured meshes the preserve information to the maximal extent is extremely challenging.

We introduce in our proposed method major contributions to address the above-mentioned challenges to form a complete fusion pipeline. These contributions are: 1) we introduce a monocular video-frame based 3D reconstruction pipeline to achieve the minimal geometric distortion by leveraging the speed and accuracy in a photogrammetric reconstruction pipeline called MetricSFM. 2) We introduce a cross-view geo-registration and fusion algorithm that takes point clouds generated from satellite multi-view stereo (MVS) images and from ground-view videos, to co-register the ground-view point clouds to the overview point clouds; 3) we extend a view-based meshing approach to accommodate point clouds with images coming from different cameras. The rest of this chapter is organized as follows: Section 2 introduces related works and the overview of the proposed pipeline; Section 3 introduces our methodologies of the components of the pipeline in details, and section 4 describes the experiment dataset and the results of the 3D reconstruction Section 5 concludes this chapter by discussing potential works moving forward.

## II. Related works and an overview of the proposed pipeline

The uses of multi-source 3D data have been attempted for different purposes, such as for localization, geo-registration, image synthesis and cartographical model generation [6-9]. For example, [8] utilized a combination of UAV (Unmanned Aerial Vehicles) images and mobile LiDAR (Light Detection and Ranging) for 3D model generation, where the geo-registrations are performed using manually measured ground control points (GCP) from the LiDAR data, followed by a Bundle Adjustment [10] of the UAV images. All were performed following a surveying-grade processes, thus minimal topographical distortions needed to be addressed in critical or non-optimally collected data (e.g., monocular video collection with a single trajectory).

Correlating the satellite overview and ground view images is extremely challenging because the areas in common can sometimes be barely the ground or even less (due to vegetations and moving objects). There are two types of approaches to address relevant tasks, such as 1) cross-view images localization [9, 11, 12] and 2) cross-view image synthesis [6, 7]. Since the traditional feature-based matching methods fail in cross-view data, the major technical approaches for cross-view data instead are to learn deep representations between cross-view data, with various strategies for learning scene-level descriptors used to match cross-view data, combing learned semantics and geometric transformation. A few early works also explored the use of manually crafted features for such a task [11, 13]. Most of the existing methods exploring 3D data co-registration, require a certain common regions and the transformation are often assumed to be simple models such as similarity or rigid transformations [14, 15]. Thus exploring methods for registering wide-area, cross-view dataset potentially with complex geometric distortions are particularly of interest and can offer tangible solutions for low-cost 3D data generation.

Meshing point clouds seems to be a standard practice with many applicable algorithms available [16]. However, for image-based point clouds, meshing requires the use of the visibility information between the view and each point [17, 18] which sometimes are not easily available for multi-source data as first of all, they may share different camera model, and second of all, standard software packages generating point clouds from images do not offer such visibility information. As a result, a standard practice of using multi-source image-based point clouds only takes point-cloud based meshing methods [16] which are designed for very dense point clouds and do not necessarily work well for point clouds with the level of uncertainty and complexity as the image-based point clouds.

Despite these challenges, we consider the problem of turning the MVS satellite images and ground-view Go-pro data to be approachable, if scenario-specific information and intermediate results of the stereo reconstruction pipeline are available. To achieve, we have the following three considerations:

1) Monocular ground-view video frames taking alongside the street do not offer an optimal camera network, thus it is possible that the results of the 3D reconstruction contain geometric distortion, for example, trajectory drifts, or topographic distortion due to the incorrectly estimated interior/exterior orientations [19], which will further add challenges to the geo-registration, we therefore consider to optimize our photogrammetric reconstruction workflow by considering self-calibration for each incremental reconstruction to minimize the potential trajectory drift.

2) We observed that in an urban environment, the boundary of objects from the satellite point clouds, e.g. buildings, might coincide well with the boundary produced by projecting the façade point clouds to the ground; therefore it can be seen as a view-invariant feature for co-registering the satellite point clouds and ground-view point clouds.

3) Meshing methods will unlikely to work well on the combined point clouds (from satellite and ground-view point clouds) without the use of visibility information. Although theoretically possible, re-implementing a meshing algorithm considering different camera models can be painstakingly trivial. We consider the satellite point clouds to associated with an orthophoto under a parallel projection, thus the visibility can be easily computed and incorporated into an image-based meshing [18] and texture mapping pipeline [20].

To sum, our proposed data generation pipeline considers three major components. As shown in Figure 1, which includes separate 3D data generation (for MVS satellite images and ground-level video frames), geo-registration and meshing.

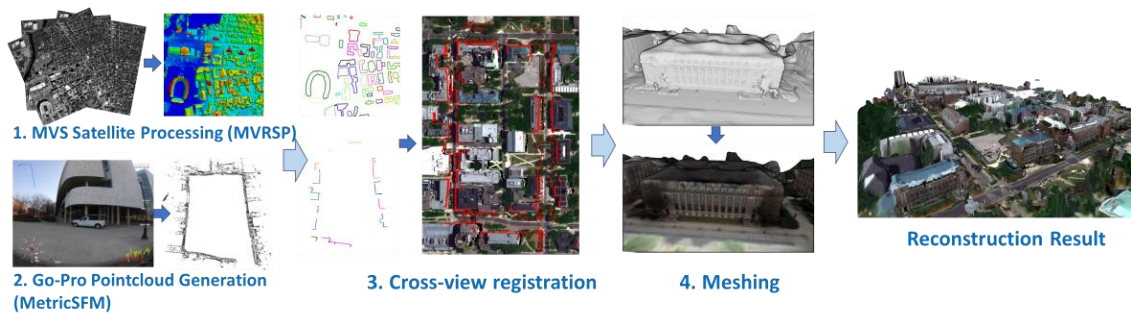

**Figure 1**. The general workflow of our processing pipeline.

The MVRSP (based on [4, 21, 22]) and MetricSFM are respectively our developed system for processing the satellite data and ground-level video frames. A cross-view registration method is performed for overview and ground-view point cloud registration, which utilize the boundary information derived from both types of point clouds. Finally, the co-registered point clouds are processed by a modified meshing and texture mapping algorithm that innovatively consider both perspective and parallelly projected image (satellite orthophoto) in an integrated optimization framework.

## III.    Methodology

### 3.1. *Multi-view (MVS) satellite image processing*

This is a pre-publication version of the published book chapter in Recent Advances in Image Restoration with Applications to Real World Problems. The final contents are subject to minor edits

The MVS satellite processing follows methods in [4, 21], which takes a pair-wise reconstruction followed by a DSM (Digital Surface Model) fusion as shown in Figure 2. Given a set of images, we will first apply an analysis algorithm presented in [23] to rank the matchability of the satellite stereo pairs (enumerated from the existing images), and then we take the top five stereo pairs to perform relative orientation and stereo dense matching using a software called RPC stereo processor [4, 22]. The core matching algorithm uses a hierarchical Semi-Global Matching [24] with modifications to accommodate large-format images [25]. The use of multiple stereo pairs enables sufficient redundancies for high-quality 3D reconstruction, and the images consist of both Worldview I/II images (data will be introduced in Section 4). The produced individual DSMs resulting from different stereo pairs are co-registered with a shift-based registration which search for translation parameters in reference to one of the pairs (which is used to be the first pair in the pair ranking), and the co-registered DSMs are fused following an adaptive depth-fusion method [21] that utilizes the color information of the orthophoto, which were shown to achieve better accuracy than a simple median depth filtering. The readers may refer to specific details of the reconstruction in [4, 21, 23].

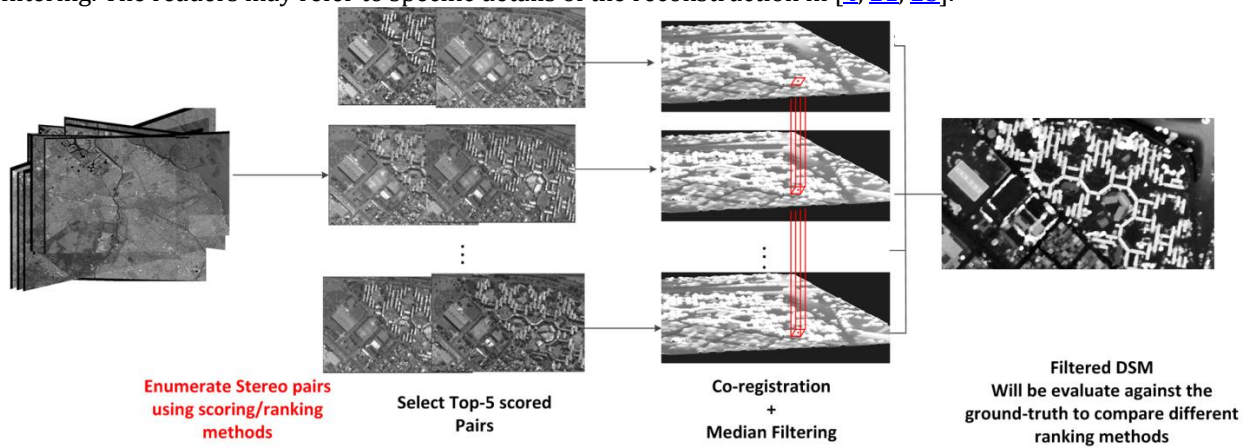

**Figure 2**. A workflow of the multi-view satellite image processing [23]

A typical Digital surface model generated using our pipeline is shown in Figure 3, which indicates a 3D reconstruction result of the central area of the city of London. Worldview-III images with a 0.3 meter resolution are used thus the resulting surface models are with the same resolution.

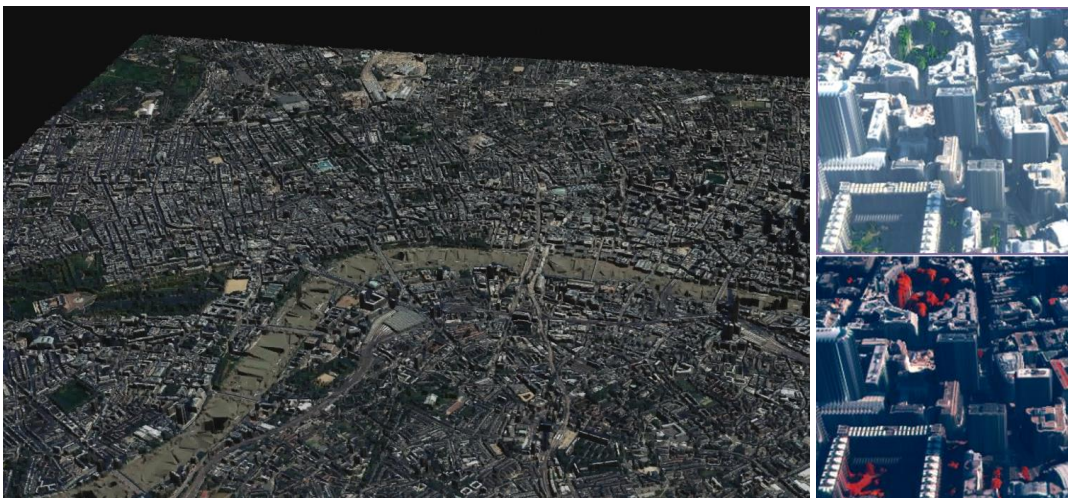

**Figure 3**. 3D reconstruction of the central area in London (ca. 50 km2). Left: overview of part of the area; right: top, enlarged RGB (Red, Green, Blue) color image, bottom, pseudo color image (near infrared, Red, Green)

3.2. *3D reconstruction from ground-view monocular image sequences*

This is a pre-publication version of the published book chapter in Recent Advances in Image Restoration with Applications to Real World Problems. The final contents are subject to minor edits

Monocular 3D reconstruction refers to the process of recovering shape of objects using images taken from a single video camera. As compared to typical stereo/multi-stereo images captured from well-distributed angles, such video sequences present sub-optimal camera network in which the pose estimation is often inaccurate for metrically correct 3D reconstruction. Oftentimes, the structure from motion and SLAM (Simultaneous Localization and Mapping) approaches are used to compute the camera poses and generate 3D semi-dense or dense point clouds. These methods although provide visually pleasant trajectories and point clouds, they may often be metrically incorrect and present drifting problems. In this section, we introduce a monocular 3D reconstruction system that leverages the speed of a typical SLAM system and rigorous photogrammetric optimization. We first present typical components for 3D reconstruction and then briefly introduce the processing workflow of the system.

### 3.2.1 A 3D reconstruction pipeline

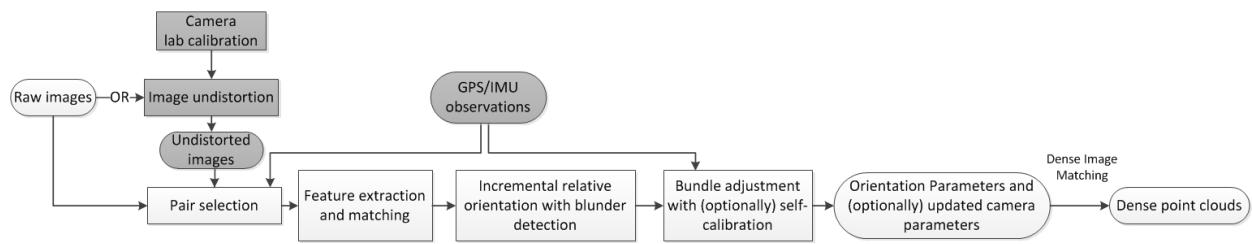

**Figure 4**. A typical 3D reconstruction pipeline, dark-grey blocks indicate optional steps.

Figure 4 presents a typical image-based 3D reconstruction pipeline. Raw images or undistorted images (through pre-calibrated parameters) are taken as the input and follow a series steps named feature extraction and matching, relative orientation, bundle adjustment and dense image matching, and output intrinsic and extrinsic orientation parameters and dense point clouds. Among these steps the GPS (global positioning System) or IMU (inertial measurement unit) can be optionally taken as observations to bring global datum. Below we briefly introduce these components and their specifics in a ground-view image sequence scenario:

**Camera intrinsic and extrinsic parameters**: The camera intrinsic parameters refer to the internal geometry of the camera and often considered as focal length, principal points and lens distortions. The extrinsic parameters refer to the poses (position and facing) of each image, normally represented by six parameters: three for a point in Euclidean coordinate (camera perspective center) and three rotation angles (sometimes are represented directly as rotation matrix).

*Pair selection*: Pair selection tells the system what are the images that are likely to observe the same object, such that a connected graph can be built [26] [27] to formulate observations to recover 3D geometry. In the ground-view scenario, this can be simply formulated using the timestamp of the frames.

**Feature extraction and matching**: features represent areas or points of interest in images and denote special pieces of information. In 3D reconstruction points are the most popular feature representations due to their simplicity and flexibility. Point features can be understood as corners or spots that are distinctive and easily identifiable across different images with various level of perspective differences, typical features are SIFT (Scale-Invariant Feature Transform) [28], SURF (Speeded up robust features) [29], and ORB (Oriented FAST and Rotated BRIEF) [30], etc. Once these points are extracted, feature matching aims to associate identical points across different images, which essentially represents corresponding rays from different images. Typically done with an exhaustive search, feature matching in a ground-view image frame scenario can be speed up by considering the fact of horizontal moving thus to reduce the search space [31].

**Incremental relative orientation/pose estimation**: The incremental relative orientation refers to the process starting with a two-view relative orientation, followed by sequentially orienting the rest of the images given the feature point correspondences. Often the estimation process needs to address blunders in the observations and the state-of-the-art procedure takes RANSAC (random sampling consensus) [32] for robust and automated relative pose estimation. RANSAC used a random sampling strategy that starts with randomly sampled feature matches (observations) instead of all the observations for relative orientation (model estimation), and runs the same process for multiple times, and select the model (estimated orientation parameters) accounting for most of the observations with reasonable residual. This has dramatically improved the automation in relative orientation and subsequently the incremental procedure, as it theoretically only requires the error rate of the matches be larger than 50%, while apparently the state-of-the-art feature extractors and matchers do much better with images in most of the applications.

**Bundle adjustment**: is a refinement process for the intrinsic and extrinsic camera parameters simultaneously with the 3D coordinated of the scene points since the measurements are prone to errors [33]. It involves a global minimization scheme using robust nonlinear least-squares algorithm such as Levenberg-Marquardt [34]. This often comes with a procedure called self-calibration [35] that simultaneous estimate the lens distortions of the camera. In a ground-view video frame scenario, because the bundle adjustment is particularly time consuming, it may sometimes be simplified to only perform local bundle adjustment instead of considering all available images.

### 3.2.2 3D reconstruction using ground-view image sequences

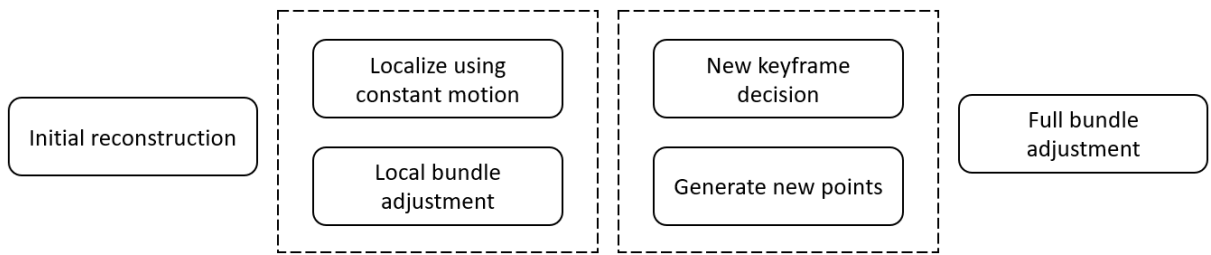

**Figure 5**. A 3D reconstruction pipeline using ground-view video frames.

Ground-view image sequences formulate a specific scenario in which a typical 3D reconstruction pipeline can be customized to accommodate the need for speed and accuracy. Our general workflow presented in Figure 5. It is similar to a SLAM pipeline [31] with the differences that the local and full bundle adjustment consider the estimation of camera lens distortion parameters. Typically, the system starts with an initialization module that aims at estimating the camera pose for the two images used in the initialization by utilizing the matched features between them, this is in line with the first half of incremental relative orientation as mentioned above. Moreover, this module generates initial 3D points of the scene by triangulating the matched feature points from the two images. After generating the initial reconstruction, the tracking module (in dashed box) starts to localize every image by finding its pose, which is similar to the second half of the relative orientation which sequentially add image frames to the system. In this module, the temporal relation between the images is used by assuming a constant velocity motion model so that we can get an initial estimate of the current image pose. Thus, using the estimated pose, we can directly project the 3D points into the current image and perform window-based search for the potential feature matches with the projected points. Consequently, we can save computations by searching correspondences only inside this window instead of searching in the whole image. Then, using these correspondences, the current camera pose can be estimated. It should be noted that the concept of keyframes are used to identify important frames in which the poses will be optimized through bundle adjustment, because frames that are estimated through a constant velocity are considered to close enough to interpolate. For images that fail the constant velocity motion model, the tracking module performs full feature matches to find feature in previous frames that have an associated map point using a spatial resection (i.e. a Perspective-n-Point (PnP) algorithm) [36] by taking existing 3D points and 2D correspondences to compute their pose, and such images are then taken as the new keyframes, in the meantime features with no 3D correspondences will be triangulated as candidates of 3D map points. Once the tracking module accumulates frames to a pre-defined number, a full bundle adjustment is used interchangeably with local bundle adjustment to refine the estimated measurements. These aforementioned processes are implemented in an in-house software package called MetricSFM. A sample from the 3D reconstruction results is shown in Figure 6.

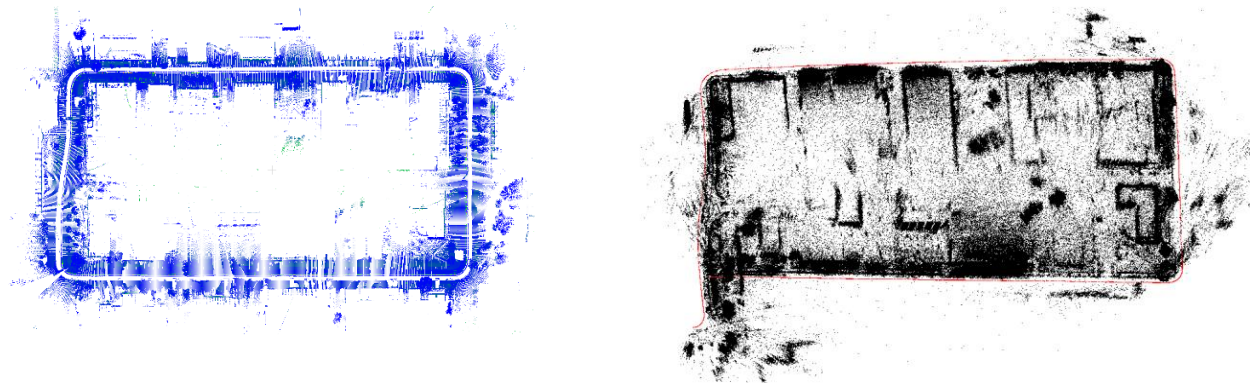

**Figure 6**. The 3D reconstruction result, left: ground truth trajectory from mobile LiDAR, right: our result without loop closure (7500 frames).

### 3.3. *Cross-view 3D point co-registration and fusion*

Non-rigid distortion of the ground-view data (e.g., trajectory drift) and very limited overlapping region among cross-view data make them difficult to be registered without significant manual effort. Based on the assumption that the object boundaries (e.g., buildings) from the over-view data should coincide with footprints of façade points from ground-view, we tackle these problems by proposing a fully automated geo-registration method for cross-view data, which utilizes semantically segmented object boundaries as view-invariant features under a global optimization framework. Taking the over-view point clouds generated from satellite stereo/multi-stereo images and the ground-view point clouds from monocular video frames as the input, the proposed method takes a "local + global" strategy to solve the non-rigid cross-view registration problem using object boundaries, which is further optimized through a constrained bundle adjustment to keep 2D-3D consistencies.

### 3.3.1 Building boundary extraction from ground-view and over-view point clouds

The building extraction on the over-view point cloud is achieved by converting the point cloud into a Digital Surface Model (DSM), on which the well-developed morphological top-hat [37, 38] can be used to extract a binary mask for all the high objects like tree and building. For satellite orthophoto containing multi-spectral information, the NDVI (Normalized Difference Vegetation Index) [39] can be extracted to further remove the trees from the binary masks. The ground-view building detection is based on the observation that the building façade points are usually vertical to the horizontal ground plane. We therefore determine the vertical direction by calculating the normal vector for all the points and then selecting the direction with the largest number of normal vectors pointing to the vertical directions. Once the vertical direction is obtained, all the ground-view points are projected onto the horizontal plane, which is followed by a classical region growing method [40] to extract point cloud segments. Finally, those segments with the number of points greater than a threshold are kept as the extracted ground-view buildings. The results of building boundary extraction from both over-view and ground-view data can be seen in Figure 7.

### 3.3.2 Local registration via building boundary matching.

In order to efficiently search for accurate registration parameters locally to address potential topographical errors of the point clouds (e.g. drifted trajectory resulting metrically incorrect point clouds), we developed a simple 2D point cloud registration algorithm that perform sampled exhaustive search. Given the over-view point set $P_d$ of size $n_d$ as the destination point cloud, and the ground-view point set $P_s$ of size $n_s$ as the source point cloud, with the scale difference $s$ between two point sets. Firstly, the distance map (as Figure 8(b) shows) for the $P_d$ is calculated via distance transformation [41, 42], in which the distance of each pixel (colored in grey-level, darkest referring to the closest distance) to the region of interest (in our scenario this refers to the boundary from the overview data). The $P_s$ is centralized by subtracting the central point for each point in $P_s$. Then we traverse the rotation space $[0, 2\pi]$ with a $3°$ interval (as an empirical parameter) for $\theta$. To speed up the computation, we group points by a certain number $c$ (e.g. 10 points) from $P_d$ and simultaneously select the most possible corresponding points from $P_s$ for each group by computing those with the closest distance, and the computation can be further speed up by building the error metrics using the pre-calculated distance map of $P_d$. The process eventually evaluates $c * n_s$ errors and concludes a translation parameter $t$ for each rotation hypothesis $\theta$ (a total of 120 given an interval of $3°$ out of $360°$). The final rotation parameter and translation parameter were found as ones that minimize the co-registration error in the distance map, and an example result is shown in Figure 8(c).

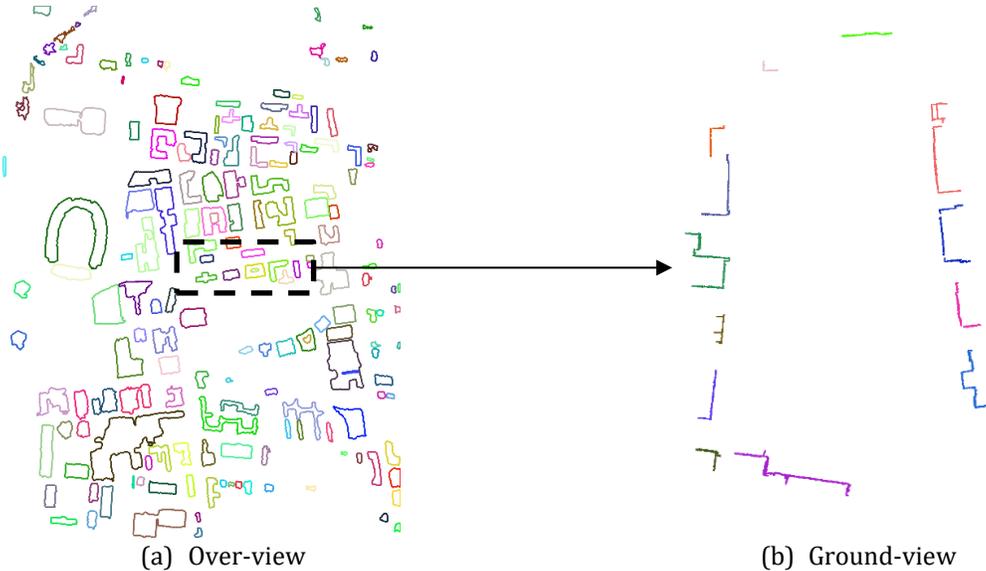

| (a) Over-view | (b) Ground-view |

**Figure 7**. Illustration of building boundary extraction results from over-view and ground-view data.

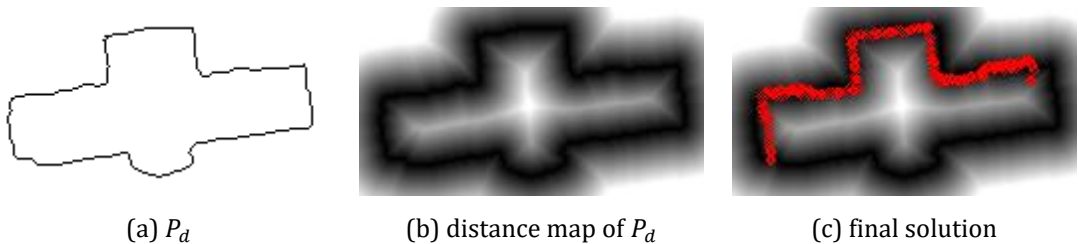

| (a) $P_d$ | (b) distance map of $P_d$ | (c) final solution |

**Figure 8**. Exhaustive search based local matching algorithm. Given the over-view building boundary points $P_d$ as destination in (a), the distance map in (b) is calculated where the intensity of pixel denotes the closest distance to $P_d$, then the global solution in (c) is obtained by our proposed method. Red points represent the ground-view points $P_s$.

Assuming there are M building segments in the overview point cloud and N building segments in the ground-view point cloud, the building matching procedure will yield M × N × 120 pair of transformation parameters ($\theta$ and $t$). Most of these are outliers as the one-to-one constraint is not yet implemented. As a quick filtering, we include additional points within a radius (we set 1/10 the length of the entire trajectory) to re-evaluate the registration errors on the distance map and discard those with average registration error larger than a threshold (e.g. 2 meters).

### 3.3.3 Global registration via energy minimization.

In the previous building segment matching step, a list of transformations $\mathcal{T} = \{T_i, i = 1,2, \dots\}$ is generated, which constitutes the final hypotheses for each building segments. We consider that the transformation hypothesis for neighboring building segments to be similar, therefore, we consider formulating this constrain in an energy minimization problem (Equation (1)):

This is a pre-publication version of the published book chapter in Recent Advances in Image Restoration with Applications to Real World Problems. The final contents are subject to minor edits

$$E(\mathcal{T}) = \sum_{B} D(B, T) + \sum_{B_i B_j} V_{B_i B_j}(T_{B_i}, T_{B_j}),$$ (1)

where $D(B, T)$ is the data term for teach building segment B with a transformation $T$ in $\mathcal{T}$, $V_{B_i B_j}(T_{B_i}, T_{B_j})$ is the smooth term that penalize differences of two transformations $T_{B_i}$ and $T_{B_j}$ of the building segments $B_i$ and $B_j$.

a) Data Term

Given a building B and a transformation $T$, we first collect its k-adjacent buildings (including B), measured using distance between barycentric coordinates. These segments after transformation are used to verify how close they are to the over-view building segments. To robustify the evaluation, we consider counting the number of points that are close enough to the overview building segments, as follows (Equation (2)):

$$D(B, T) = \sum_{p \in B} c(p, p') = \begin{cases} 0, & \text{if } d(p, p') < d_{th} \\ 1, & \text{otherwise} \end{cases}$$ (2)

where $c(p, p')$ is the cost of a point p that belongs to the building B, which equals to 0 if the distance $d(p, p')$ between p and its closest point p' in the over-view building boundaries is smaller than $d_{th}$, and equals to 1 otherwise. This formulation can effectively keep the value range of the data term limited. For example, the value of $d(p, p')$ can be very large if an incorrect transformation converts the point p far away from p', however $c(p, p')$ can eliminate the influence of this mistake to generate more reasonable cost value.

b) Smooth Term

The smooth term $V_{B_i B_j}(T_{B_i}, T_{B_j})$ penalizes the transformation associate with two neighboring buildings being too different, shown in Equation (3):

$$V_{B_i B_j}(T_{B_i}, T_{B_j})$$
$$= \begin{cases} p1, \text{if } \left\| \theta_{B_i} - \theta_{B_j} \right\| < \theta_{th} \text{ and } \left\| t_{B_i} - t_{B_j} \right\| < t_{th} \\ p2, \text{otherwise} \end{cases}$$ (3)

where $\theta$ is the rotation angle in 2D and $t$ is translation, and we have $T = (s, R, t)$ with $R = \begin{pmatrix} \cos(\theta) & -\sin(\theta) \\ \sin(\theta) & \cos(\theta) \end{pmatrix}$, $p1$ gives a smaller penalty for assigning two similar transformations $T_{B_i}$ and $T_{B_j}$ with $\left\| \theta_{B_i} - \theta_{B_j} \right\| < \theta_{th}$ and $\left\| t_{B_i} - t_{B_j} \right\| < t_{th}$ to two adjacent buildings $B_i$ and $B_j$, $p2$ is the larger penalty. The values of $p1$ and $p2$ are set in Equation (4):

$$p1 = 2h$$
$$p2 = \left( \frac{\left\| \theta_{B_i} - \theta_{B_j} \right\|}{\theta_{th}} + \frac{\left\| t_{B_i} - t_{B_j} \right\|}{t_{th}} \right) * h$$ (4)

where $h$ is a constant that converts the smooth term into the same magnitude with the data term, which is defined as the $1/100$ the total number of ground-view building boundary points, $\theta_{th}$ is the angle threshold which is set as $10°$, and $t_{th}$ is the translation threshold which is set as 100 meters. The solution Equation (1) can be achieved efficiently through graph-cut algorithm [43].

3.3.4 Bundle adjustment for pose refinement

The co-registration is further performed in the vertical direction using the overlapping ground points, and this is followed by a bundle adjustment of all image poses such that they are consistent with the registered ground-view point clouds. This is achieved by weighting the unknown poses to be close to the poses after the transformation. An additional bundle adjustment benefits the poses to be strictly following the epipolar constraints thus offers consistent 2D-3D relationship for further processing such as texture mapping. Both the overview and ground-view point clouds are then combined and their overlapping point clouds were fused as follows: for areas where both satellite point clouds and ground-view point clouds exist, we take the ground-view point clouds as it with a resolution presents higher accuracy and certainty. An example of co-registered cross-view point clouds is shown in Figure 9.

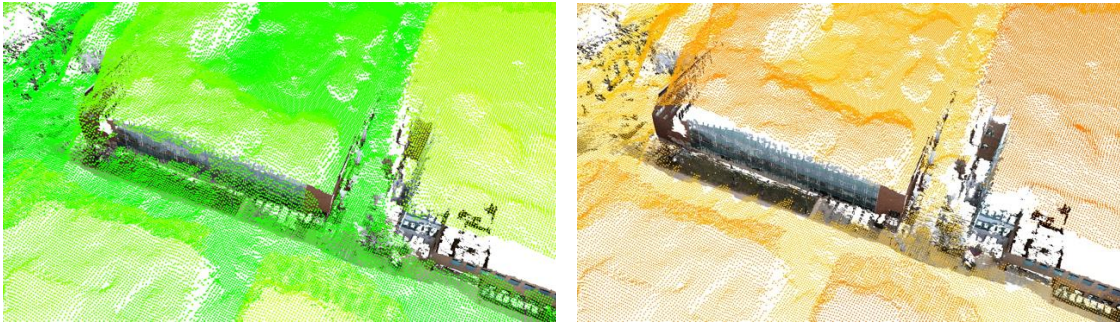

**Figure 9**. Co-registered cross-view point clouds are fused (left: before, right: after) by only keeping the high resolution results. Non-textured points are over-view satellite point clouds.

### 3.4. Meshing and texture mapping of cross-view fused point clouds

### 3.4.1 Mesh reconstruction of cross-view fused point cloud

As mentioned in Section I (Introduction), a point cloud based meshing method [16] is unlikely to yield visually consistent meshes (an example is shown in Figure14 ). Therefore, our solution considers the use of image information for mesh reconstruction. The base method [18] takes the constructed Delaunay tetrahedra of the point clouds as the input to extract the surface. These tetrahedra can be viewed as a connected graph, in which the tetrahedra are the nodes and shared/common faces are edges. Figure 10 shows the procedure: black triangles denote cameras, dash arrows denote visual rays, each point in 3D space can be determined by at least two rays, which connect the object points and camera centers, here we call it ray visibility. Based on ray visibility, tetrahedra intersected with rays are evaluated by their probability to be in a free space (outer space), and tetrahedra behind the ray endpoint are evaluated by their probability belonging to the full space (inner space). Such a graph labeling can be casted to a s-t minimal cut problem and solved with maxflow algorithms [44]. The final surfaces are the common faces of the tetrahedra labeled as free and full spaces (Figure 10).

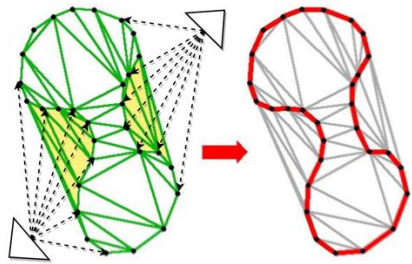

**Figure 10**. Left: Green network is Delaunay triangulation, yellow region (free space) are tetrahedra which intersected with rays (dash arrows), white region is tetrahedra labelled as full space. Right: Red lines are surfaces between full and free space, which are common faces shared by those tetrahedra (artwork from [45] with minor edit).

Our pipeline extends from this base algorithm by incorporating point clouds generated from the satellite images. The following steps give streamline from source points to surface mesh model.

**<u>Delaunay 3D Triangulation:</u>** 3D Triangulation or tetrahedralization is extended from 2D triangulation, which partitions a polyhedron into non-overlapping basic 3D elements, where the vertices of tetrahedra take the vertices of the original polyhedron. Delaunay tetrahedron reconstruction [46] divides the convex hull of points into compact simplices, where neither extremely long edge nor extremely sharp angle is included.

Many well-known commercial packages and open source projects have implemented the algorithm that creates Delaunay tetrahedron from point set, here we use CGAL [47] an open source computational geometric algorithm library to construct tetrahedra.

**Visibility:** Each ray will propagate its confidence to intersected nodes (tetrahedra) and edges (triangle faces) of the tetrahedra graph. The algorithm was implemented by an open-sourced project OpenMVS [48]. Dense points and their associated images with poses are the most common source of visibility in our framework, often under a perspective geometry. However, the geometric model of satellite camera sensors is different (e.g. Rational Polynomial Coefficients) [4]. By considering that the point clouds can be associated with the orthophoto through a parallel projection, we proposed a two-step method: 1) project satellite point on to grid, only the highest point is recorded in each cell. 2) Create vertical visual rays from those points.

**Assigning Weights for the Graph:** Our method follows a so-called soft visibility weighting model that was used by the base algorithm. The readers may refer to the original paper [18] for more detail.

**Solving Min-Cut problem:** Once weighting procedure for the edges is done, we use IBFS (Incremental Breadth First Search) [49] maximum flow algorithm to solve minimum *s-t* cut problem. And finally, the common faces between source and sink tetrahedra are extracted to build up optimum surface model.

3.4.2 Texture mapping of cross-view fused point cloud

Our texture mapping framework is based on Waechter 's work [20] which has been well practiced and widely used by rather popular open source projects, e.g. OpenMVS [48]. Texturing a 3D model from multiple registered images is typically performed in a two steps approach: 1) select view(s) should be used to texture each face yielding a preliminary texture, and 2) optimize the texture to avoid seams between adjacent texture patches.

**Best View Selection:** The base method [20] determines face visibility (distinct from ray visibility) for all combinations of views and faces by first performing back face and view frustum culling, then renders faces onto images, using depth buffer to determine the nearest faces. Lempitsky and Ivanov [50] [51] [52] compute a labeling that assign a view to be used as texture for each mesh face using a pairwise Markov random field energy formulation. We consider the ground-view images are perspective and the satellite orthophotos are in parallel projection. Our texture mapping considers the orthophoto as one of the images with only few simple modifications: we balanced data term of ortho images to compensate resolution gap and make ortho images as the default sources for texturing.

**Seamless Texture Fusion:** In Waechter et al.'s method [20], they a global and local color adjustment method to blur the seams, which extended Lempitsky and Ivanov's [50] color adjustment approach. The original approach only accounts for color difference on vertices to measure color difference along the seam line, called global adjustment. The extended method added a local adjustment with Poisson editing [52] affect border strip of image patches. In our case, since the resolution of orthophoto is way lower than the ground-view images, prior to applying the fusion of image patches, we up-sampled orthophoto to the same resolution as that of the ground-view images. After color balancing and Poisson editing, color differences can be well-adjusted and seams are successfully been blurred.

## IV.  Data Description

We take the Ohio State University (OSU) Columbus Campus as our test site, of which we have collected twelve overlapping satellite images consisting of WorldView-I and WorldView-II images (information shown in Table. 1). These images selectively form 31 pairs used for the reconstruction based on the method of [23], and many of these images are not from the same year thus creating challenges for the reconstruction. Table 2. provides an overview of the first 10 pairs used from the acquired images: not all of these pairs forms in-track stereo, while the large redundancy does provide the advantage in producing more accurate surface model. Figure 11. shows the generated digital surface model. The achieved RMSE (root-mean-squared-error) is 1.26 meters evaluated through LiDAR point clouds, and the RMSE reached 0.60 meters by excluding changed buildings, rivers and trees.

This is a pre-publication version of the published book chapter in Recent Advances in Image Restoration with Applications to Real World Problems. The final contents are subject to minor edits

**Table 1**. Twelve overlapping satellite images used for satellite-based 3D reconstruction.

| | Acquisition time | Sensor | Off nadir(degree) | Sun elevation angle (degree) | Resolution (meter) | Cloud cover percentage (%) |
|---|---|---|---|---|---|---|
| 1 | 2009-04-01 | WorldView-01 | 1.80 | 52.40 | 0.50 | 0.00 |
| 2 | 2010-04-15 | WorldView-01 | 15.40 | 58.20 | 0.52 | 0.00 |
| 3 | 2010-09-25 | WorldView-02 | 13.00 | 48.30 | 0.49 | 0.04 |
| 4 | 2010-09-25 | WorldView-02 | 19.20 | 48.30 | 0.52 | 0.01 |
| 5 | 2011-10-08 | WorldView-02 | 4.30 | 43.80 | 0.47 | 0.00 |
| 6 | 2012-01-09 | WorldView-01 | 20.00 | 26.10 | 0.55 | 0.00 |
| 7 | 2012-01-09 | WorldView-01 | 32.70 | 26.20 | 0.67 | 0.00 |
| 8 | 2013-08-06 | WorldView-02 | 15.80 | 64.20 | 0.50 | 0.00 |
| 9 | 2013-12-28 | WorldView-01 | 22.90 | 24.50 | 0.57 | 0.00 |
| 10 | 2014-06-06 | WorldView-02 | 23.50 | 70.80 | 0.54 | 0.00 |
| 11 | 2015-04-17 | WorldView-02 | 25.60 | 56.80 | 0.56 | 0.00 |
| 12 | 2019-01-05 | WorldView-02 | 19.90 | 26.60 | 0.52 | 0.00 |

**Table 2**. Examples of Metadata of pairs used for satellite-based 3D reconstruction. These data come in level 1. The Image ID refers to those in Table 1.

| Pair | Intersection angle (degree) | Sun difference angle (degree) | Time difference (days) | Left image ID | Right image ID |
|---|---|---|---|---|---|
| 1 | 6.20 | 0.00 | 0 | 3 | 4 |
| 2 | 12.70 | 0.10 | 0 | 6 | 7 |
| 3 | 13.60 | 5.80 | 379 | 1 | 2 |
| 4 | 2.90 | 1.60 | 719 | 6 | 9 |
| 5 | 9.80 | 1.70 | 719 | 7 | 9 |
| 6 | 8.70 | 4.50 | 378 | 3 | 5 |
| 7 | 14.90 | 4.50 | 378 | 4 | 5 |
| 8 | 7.70 | 6.60 | 304 | 8 | 10 |
| 9 | 2.10 | 14.00 | 315 | 10 | 11 |
| 10 | 5.70 | 30.20 | 1359 | 11 | 12 |

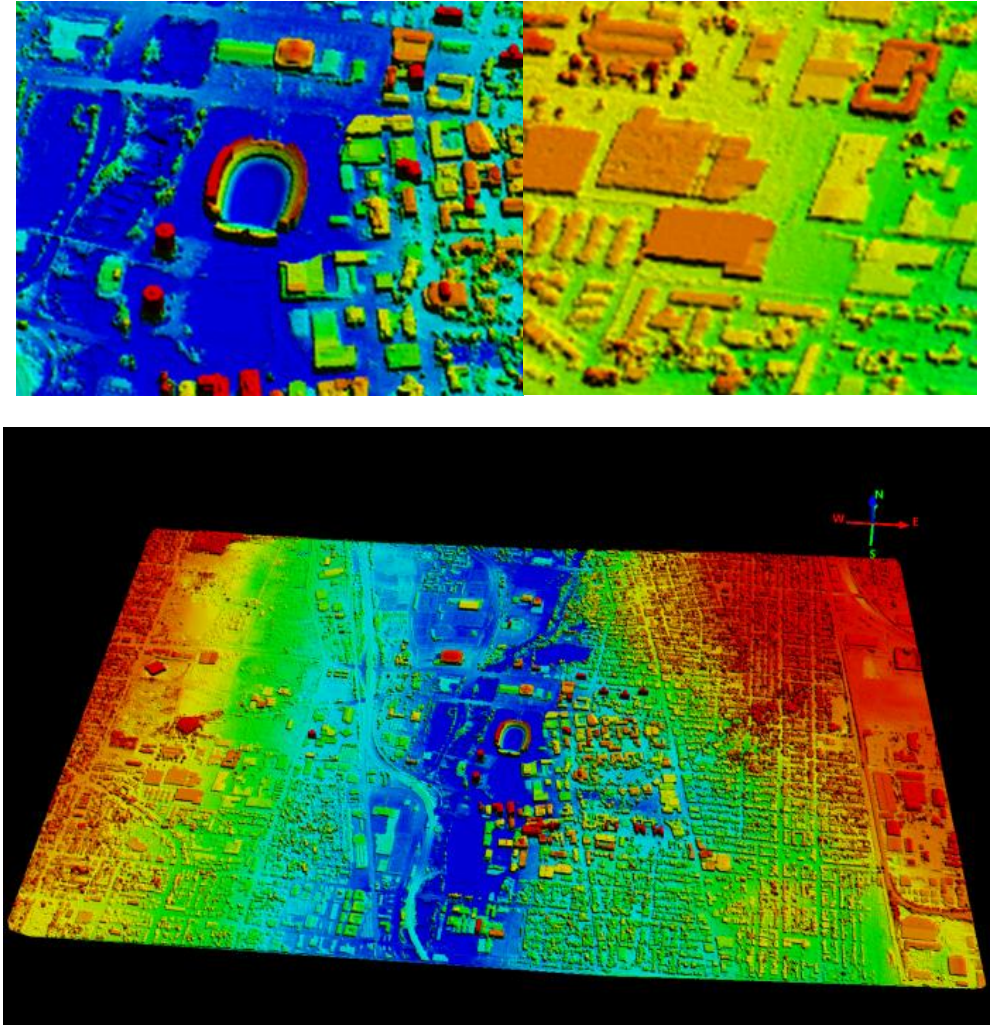

**Figure 11**. The generated Digital Surface Models of the OSU campus using our satellite data processing pipeline. The top-row shows enlarged views.

We have also collected approximately 300 GB of Go-pro videos covering a trajectory equivalent to 33 km, and the reconstruction for the ground-view images take 150k frames (with a resolution of 1500 × 2000 pixels per frame) out of these videos. Figure 12. shows the reconstructed point clouds of approximately two thirds of the region. The pose estimation time takes approximately 20 hours and dense matching takes 4hrs in a normal i-7 desktop computer.

## I.    Experiment Results

We demonstrate that the resulting geometry shows completeness in terms of the rooftop and façade information (for places where ground-view images are available). Figure 13 provides an overview of the registered point clouds and a comparison showing the mis-registration using a typical point cloud based algorithm [15]. Our co-registration achieves an RMSE of 1.44 m in error, which are reasonable considering that the satellite point clouds have a resolution of 0.5 m.

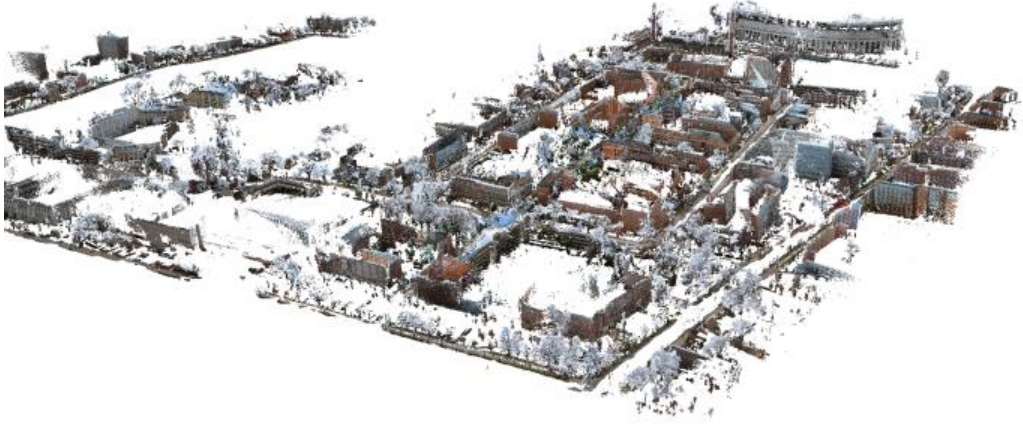

**Figure 12**. Dense reconstruction using our processing pipeline for two thirds of the campus region, totaling 7 billion color points.

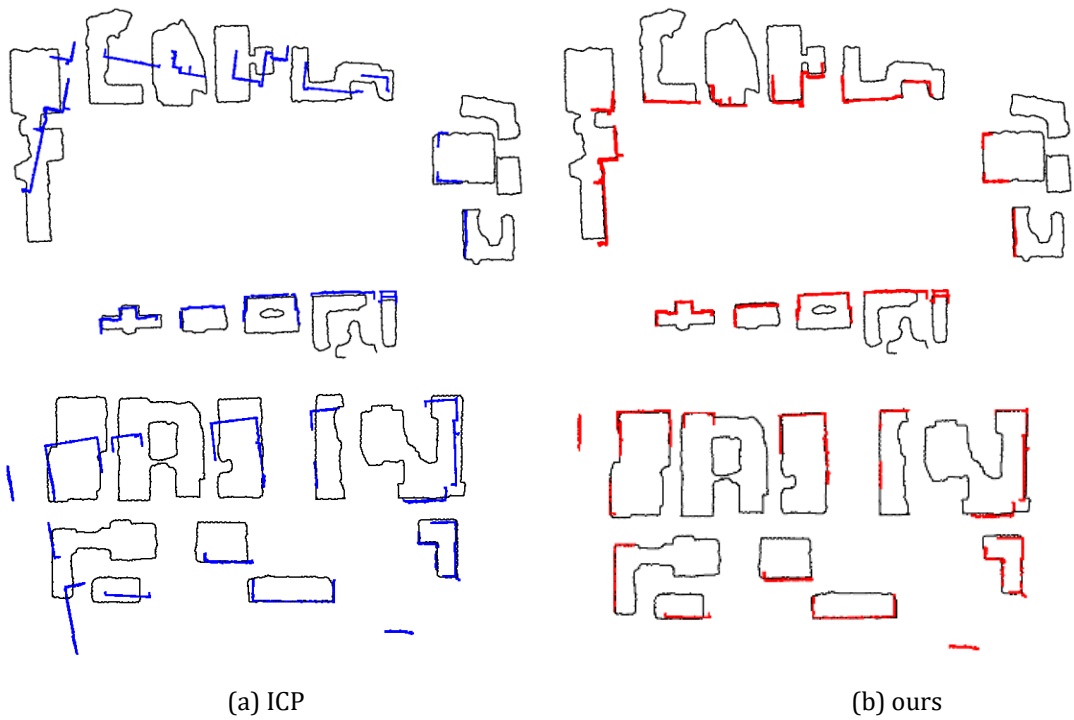

(a) ICP                                          (b) ours

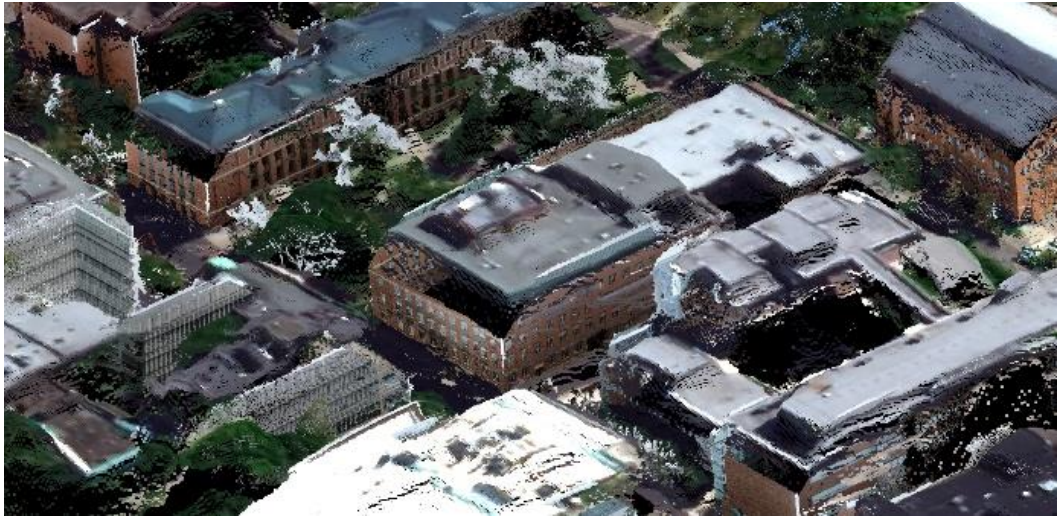

(c) Co-registered point clouds, point size of the satellite point clouds and ground-view point clouds are appropriately adjusted to optimize the visualization in this figure.

**Figure 13**. Registration result of ICP (a) and our method (b) on the distorted ground-view trajectory. (c) shows part of the registered ground-view point clouds generated on 150k Go-Pro images.

With the registered point clouds, we can generate the meshes using our proposed meshing pipeline introduced in section 3.4. Figure 14 shows the reconstructed meshes (shaded and textured) using our pipeline, and we have also included the results from a pure point cloud based meshing method, which visually demonstrates much worse results. In Figure 15, we have also included the reconstruction results of a relatively larger region using our reconstructed pipeline.

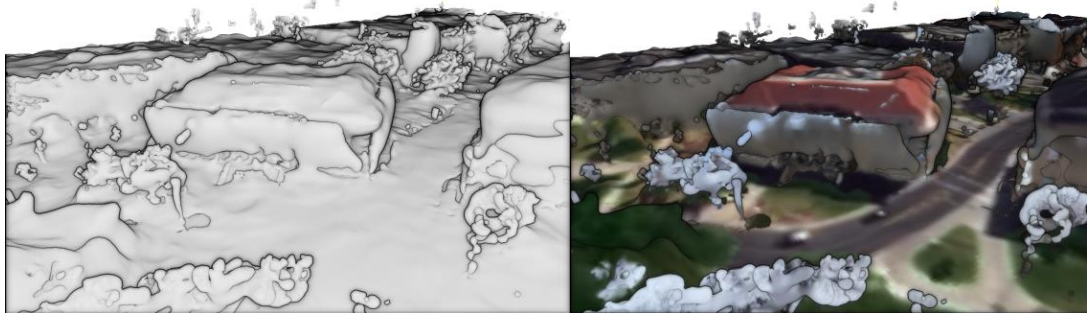

(a) Reconstructed mesh using Poisson reconstruction

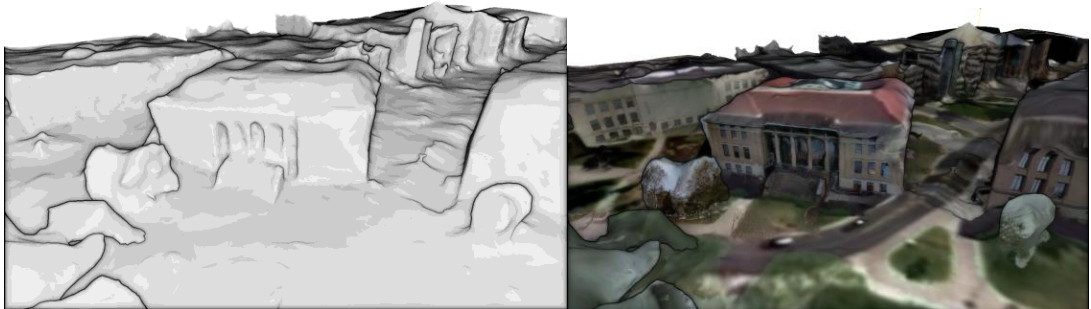

(b) Reconstructed mesh using our reconstruction method
**Figure 14**. Left: shaded mesh model. Right: textured mesh model.

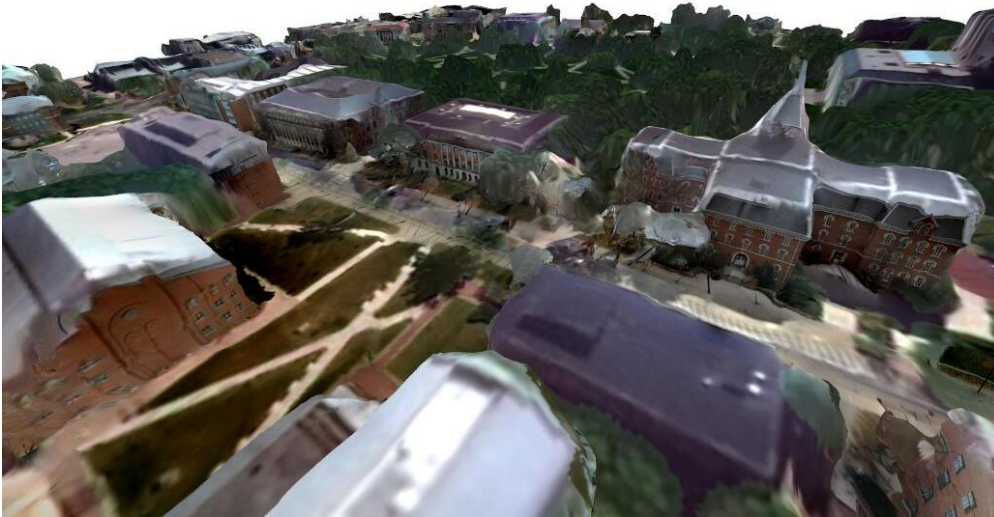

**Figure 15.** A screenshot of the generated textured mesh of the OSU campus area using our proposed pipeline, which includes information from the top-view and details on the facades.

### *Accuracy Evaluation:*

We have compared the resulting combined model with the ground truth Airborne LiDAR data as shown in Figure 16, in which we include two sample areas (top and bottom row of Figure 16). Since the airborne LiDAR does not cover the façade information, we evaluate the accuracy of the results using resampled DSM to the same grid. It is expected that the combined model with the incorporated street-view point clouds should have better accuracy given the more accurate point clouds of the (partial) ground and building boundaries. From Figure 16, we can observe that the satellite DSM (left column), due to the lower resolution, shows blurred object boundaries, as compared to the combined model (middle column). Figure 17 plots the error distributions and it evidences our observations in Figure 16: the object boundaries in the satellite DSM shows larger errors than the combined model, and it can be also seen in some regions of the ground that the combined model presents less error due to the captured fine ground structures (marked in red circle of Figure 16 and 17, bottom row). Table 3 calculates the RMSE (root mean squared error) of these two areas, and it shows that the combined model improves at 0.20 meter in accuracy for area 1 and 1 meter for area 2. This shows significant improvement in terms data accuracy, and we should note that this evaluate is only on the DSM and it is expected that if the façade data evaluation are considered (if ground truth of the façade geometry is available), the accuracy improvement can be significantly more.

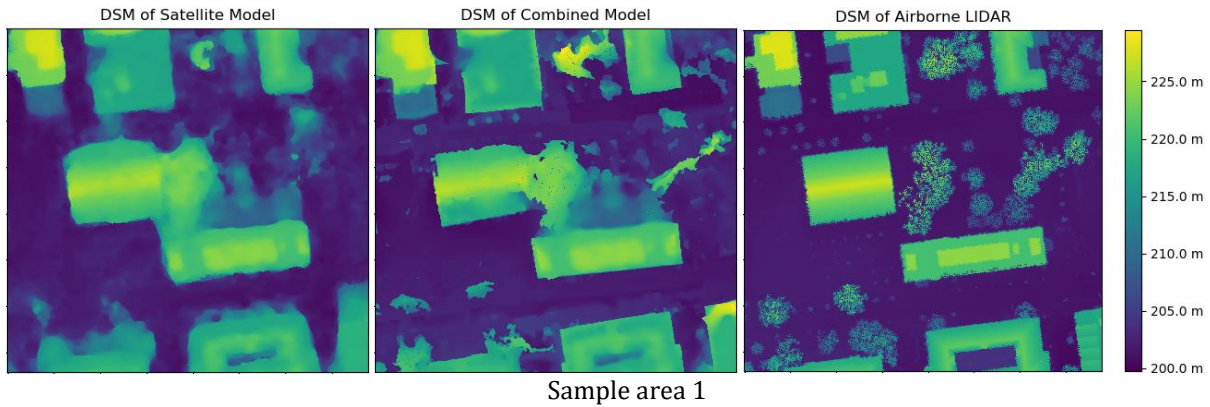

Sample area 1

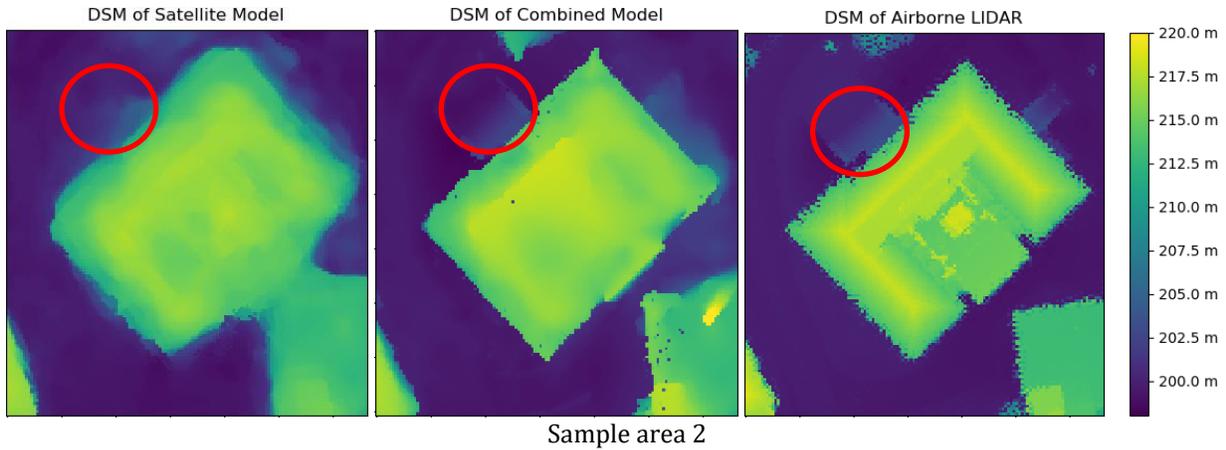

Sample area 2

**Figure 16**. DSM from satellite stereo (left column) / combined model (middle column) / airborne LIDAR (right column). Top and bottom row indicate two difference samples (sample area 1 and sample area 2). The red-circled region show that a ground structure is well compared in the combined model, as compared to the satellite DSM.

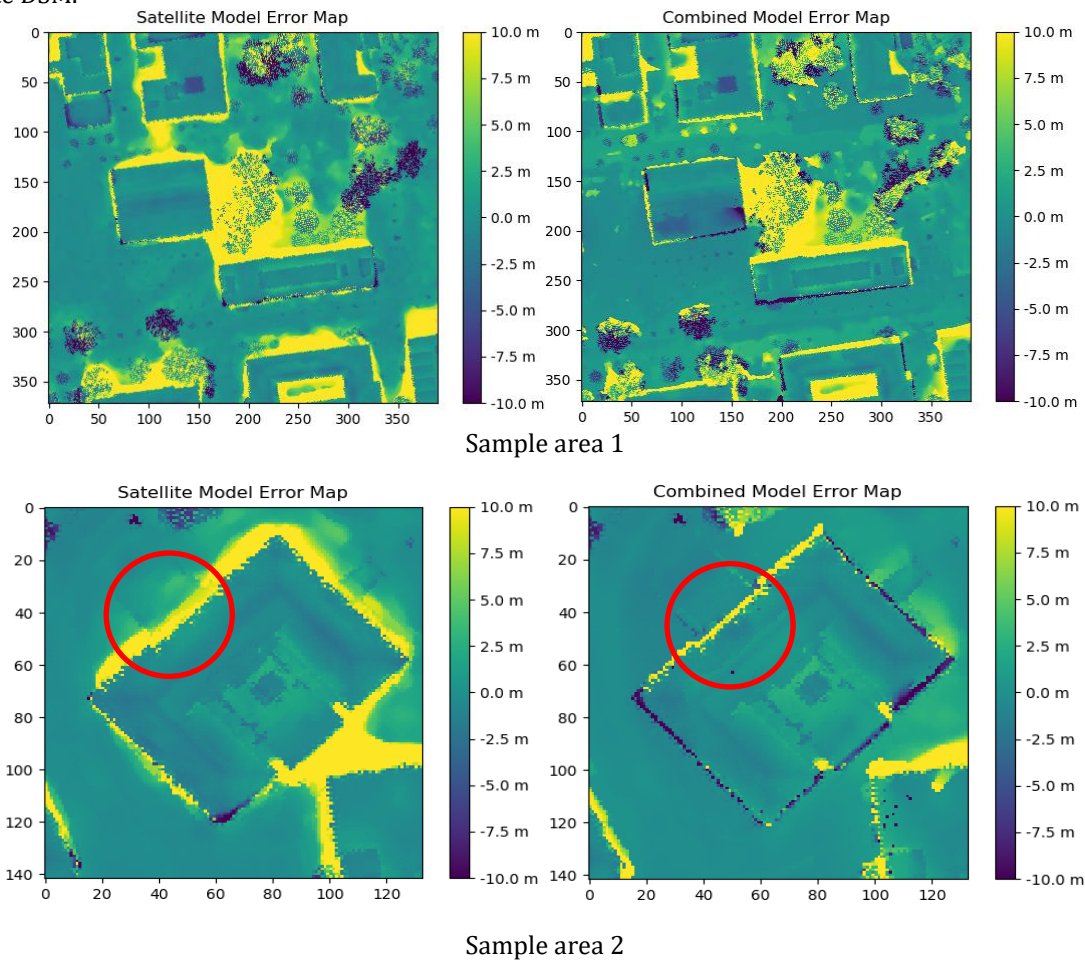

Sample area 1

Sample area 2

**Figure 17**. Error maps of satellite model (left column) and combined model (right column) evaluated against the LiDAR DSM. Top and bottom row indicate two difference samples (Sample area 1 and sample area 2). The red circled region shows smaller errors in the combined model due to that the ground structure are well captured.

Table 3. Error evaluation

|  | RMSE (m) – Area 1 | RMSE (m) – Area 2 |
|---|---|---|
| Satellite Model | 4.315 | 3.505 |
| Combined Model | 4.138 | 2.532 |

## II.      Conclusion

In this chapter, we propose a framework for fusing results from cross-view images for 3D mesh reconstruction. We present our processing framework (Figure 1.) that consists of three major components: 1) 3D reconstruction separately from the top-view satellite images and ground-level images; 2) Cross-view geo-registration between the satellite point clouds and ground-view point clouds; 3) Meshing reconstruction based on the combined satellite and ground point clouds. In each of these components, we present our developed systems and on-going research efforts in addressing the potential challenges (introduced in Section 1.1), and the in-progress results. We demonstrate that our proposed pipeline can achieve visually more consistent textured meshes, in comparison to a standard and intuitive processing method. The proposed framework and the attempts for integrating satellite and ground-view images and converting them to textured models can be of particular interest for data collection in areas where standard datasets such as aerial/UAV (unmanned aerial vehicle) photogrammetric/LiDAR flights. We have demonstrated that DSM generated from the combined model using our workflow, can be one-meter more accurate than the satellite DSM, and is expected to be much more accurate if the evaluation on the façade is considered (as the satellite DSM does not have façade information at all). Our future works include further optimizing individual modules of our processing pipeline and part of these modules will be made available once they are optimized for practical uses.

## III.      Acknowledgement

This work is supported by the Office of Naval Research (Award No. N000141712928). The satellite datasets are provided by Digital-Globe.  The authors appreciate the helpful support of Mr. Xiaohu Lu and Dr. Xu Huang in their prior work.

This is a pre-publication version of the published book chapter in Recent Advances in Image Restoration with Applications to Real World Problems. The final contents are subject to minor edits

This is a pre-publication version of the published book chapter in Recent Advances in Image Restoration with Applications to Real World Problems. The final contents are subject to minor edits